\documentclass[twoside]{article}

\usepackage{PRIMEarxiv}

\usepackage[utf8]{inputenc} 
\usepackage[T1]{fontenc}    
\usepackage{hyperref}       
\usepackage{url}            
\usepackage{booktabs}       
\usepackage{amsfonts}       
\usepackage{nicefrac}       
\usepackage{microtype}      
\usepackage{lipsum}
\usepackage{fancyhdr}       
\usepackage{graphicx}       

\usepackage[figuresright]{rotating}
\usepackage{csquotes}
\usepackage{hyperref}
\usepackage{url}
\usepackage[nolist]{acronym}

\title{Explainable Artificial Intelligence (XAI) 2.0: \\ A Manifesto of Open Challenges and Interdisciplinary Research Directions}

\author{
Luca Longo\\ The Artificial Intelligence and Cognitive Load Research Lab, Technological University Dublin, Ireland \\ \texttt{Corresponding author: luca.longo@tudublin.ie} 
\And
Mario Brcic\\ University of Zagreb, Faculty of Electrical Engineering and Computing, Croatia 
\And
Federico Cabitza\\ University of Milano-Bicocca, Milan, Italy, IRCCS Ospedale Galeazzi Sant'Ambrogio, Milan, Italy 
\And
Jaesik Choi\\ Kim Jaechul Graduate School of AI, Korea Advanced Institute of Science \& Technology, Republic of Korea\\ INEEJI Corporation, Republic of Korea
\And
Roberto Confalonieri\\ Department of Mathematics, University of Padua, Italy 
\And
Javier Del Ser\\ TECNALIA, Basque Research \& Technology Alliance (BRTA), Derio, Spain, \\ University of the Basque Country (UPV/EHU), Bilbao, Spain
\And 
Riccardo Guidotti\\  \hspace{3cm} University of Pisa, Pisa, Italy \ \hspace{3cm}
\And
Yoichi Hayashi\\ Department of Computer Science, Meiji University, Tokyo, Japan 
\And
Francisco Herrera\\ Department of Computer Science and Artificial Intelligence, \\ DaSCI Andalusian Institute in Data Science \& Computational Intelligence, University of Granada, Granada, Spain 
\And
Andreas Holzinger\\ Human-Centered AI Lab, University of Natural Resources and Life Sciences Vienna, Austria 
\And
Richard Jiang\\ School of Computing and Communications, Lancaster University, UK 
\And
Hassan Khosravi\\ The University of Queensland, Brisbane, Australia 
\And
Freddy Lecue\\ National Institute for Research in Digital Science and Technology (INRIA), Sophia Antipolis, France
\And
Gianclaudio Malgieri\\ eLaw Center for Law and Digital Technologies, Leiden University, Netherlands 
\And
Andrés Páez\\ Department of Philosophy, Universidad de los Andes, Bogotá, Colombia \\ Center for Research \& Formation in Artificial Intelligence, Universidad de los Andes, Bogotá, Colombia 
\And
Wojciech Samek\\ Technical University of Berlin, Berlin, Germany, \\ Fraunhofer Heinrich Hertz Institute, Berlin, Germany, \\ Berlin Institute for the Foundations of Learning and Data (BIFOLD), Berlin, Germany 
\And
Johannes Schneider\\ Department of Information Systems and Computer Science, University of Liechtenstein, Liechtenstein 
\And
Timo Speith\\ Department of Philosophy, University of Bayreuth, Bayreuth, Germany\\ Center for Perspicuous Computing, Saarland University, Saarbr\"{u}cken, Germany 
\And
Simone Stumpf\\School of Computing Science, University of Glasgow, United Kingdom
}

\begin{document}
\maketitle

\begin{abstract}
As systems based on opaque \ac{AI} continue to flourish in diverse real-world applications, understanding these black box models has become paramount. 
In response, \ac{XAI} has emerged as a field of research with practical and ethical benefits across various domains.
This paper not only highlights the advancements in \ac{XAI} and its application in real-world scenarios but also addresses the ongoing challenges within \ac{XAI}, emphasizing the need for broader perspectives and collaborative efforts. 
We bring together experts from diverse fields to identify open problems, striving to synchronize research agendas and accelerate \ac{XAI} in practical applications. 
By fostering collaborative discussion and interdisciplinary cooperation, we aim to propel \ac{XAI} forward, contributing to its continued success. 
Our goal is to put forward a comprehensive proposal for advancing \ac{XAI}. 
To achieve this goal, we present a manifesto of 27 open problems categorized into nine categories. 
These challenges encapsulate the complexities and nuances of \ac{XAI} and offer a road map for future research. For each problem, we provide promising research directions in the hope of harnessing the collective intelligence of interested stakeholders.
\end{abstract}


\begin{acronym}
\acro{AI}{Artificial Intelligence}
\acro{AIED}{\acs{AI} in Education}
\acro{DL}{Deep Learning}
\acro{DT}{Decision Tree}
\acro{DNN}{Deep Neural Network}
\acro{HLEGAI}{High-Level Expert Group on \acs{AI}}
\acro{LIME}{Local Interpretable Model-Agnostic Explanations}
\acro{LLM}{Large Language Model}
\acro{LRP}{Layer-Wise Relevance Propagation}
\acro{ML}{Machine Learning}
\acro{SHAP}{Shapley Additive Explanation}
\acro{XAI}{Explainable \acs{AI}}
\end{acronym}

\section{Introduction}
The field of \acf{XAI} has undergone significant growth and development over the past few years. 
It has evolved from being a niche research topic within the larger field of \acf{AI} \cite{swartout1991explanations, paris1991generation, WIREs-2020} to becoming a highly active field of research, with a large number of theoretical contributions, empirical studies, and reviews being proposed every year \cite{Speith2022Review, tintarev_survey_2007, Chakraborty2017Interpretability, Biran2017Explanation, AdadiB18, Guidotti2018, carvalho2019mac, Markus2021Role, VILONE202189, Zhou2021Evaluating, zini2022explainability, tiddi2022knowledge, mei2022explainable, Langer2021What}. 
Furthermore, \ac{XAI} has evolved into an exceedingly multidisciplinary, interdisciplinary, and transdisciplinary field. 
Among others, \ac{XAI} is now a research topic in a broad range of disciplines outside of computer science, such as engineering, chemistry, biology, education, psychology, neuroscience, and philosophy \cite{Langer2021What, Langer2021Auditing, doshi2017towards, ali2023explainable}.
The growth of \ac{XAI} can be attributed to the increasing success of \ac{AI}. 
In recent years, \ac{ML}, specifically \ac{DL}, has been successfully used in many real-world applications due to its ability to learn and automatically extract patterns from complex and non-linear data. 
\ac{ML} and \ac{DL} techniques have been used for classification, forecasting, prediction, recommendation, and data generation. 
The success of these techniques and their application in critical areas such as finance \cite{Longbing2022AI} and healthcare \cite{caruana_intelligible_2015} has made it necessary to understand these models' underlying mechanisms and their often opaque outputs. 
\ac{XAI} has emerged as a response to this demand, as it seeks to develop methods for explaining the behaviour and outputs of \ac{AI} systems. 
In other words, \ac{AI}'s need for transparency and interpretability has made \ac{XAI} an area of study with practical and ethical value in various fields \cite{Speith2022Review, Langer2021Auditing}. 
Despite much progress in \ac{XAI}, many open problems require further exploration.
For instance, \ac{XAI} alone is not enough for trustworthiness, and there is a lack of consensus concerning the priorities and directions needed to advance this research field. 
Often these open problems are viewed through siloed perspectives \cite{Miller2017Explainable, Speith2022Review}. A broader, multidisciplinary approach that draws on the expertise of researchers across different fields could bring about advances towards \ac{XAI} 2.0.
Our work addresses this gap by bringing together a wide range of experts from diverse fields to collaborate on identifying and tackling open problems in \ac{XAI}. 
The focus is on synchronizing the research agendas of scholars working in the field, identifying directions that could catalyze \ac{XAI} in real-world applications.

The goal is to form a proposal open to discussion, a sincere attempt at fostering a debate around \ac{XAI} and what research should be addressed in the future. 
By doing so, we hope to offer new insights and perspectives on, for instance, developing and improving methods and applying existing methods in novel domains.
Through this collaborative effort, we seek to advance the field of \ac{XAI} and contribute to its continued growth and success. 
In particular, we seek to propose a manifesto that comprises several propositions governing scientific research in the field of \acf{XAI}.
To achieve the goal described above, this article has come about through a very specific synthesis. 
To get different perspectives on \ac{XAI} various experts from different disciplines, including philosophy, psychology, HCI, and, of course, computer science were brought together. 
This significant effort has resulted in a total of 27 problems with their challenges, which we have divided into nine categories of two to four problems.

Overall, the structure of this paper is as follows. Through the illustration of some of the many possible use-cases of \ac{AI}, \ref{sec:xaiResearch} presents a variety of advances of \ac{XAI} techniques and methods, along with their application in real-world settings. 
This is meant to highlight the benefits of XAI to people, businesses, institutions, and society.
Subsequently, the article's core follows in \ref{sec:challengesResearchDirections} by describing 27 problems in XAI, the challenges in solving them, and our suggestions for possible solutions. 
Finally, \ref{sec:novel Manifesto} summarizes our manifesto concisely, offering a roadmap for future research.

\section{Advances and Applications of xAI Research}
\label{sec:xaiResearch}
In this section, we showcase that research in \ac{XAI} is alive and useful. 
In particular, \ref{sec:breakthroughs} focuses on synthesizing the main breakthrough in \ac{XAI}, demonstrating its enormous potential. 
Similarly, \ref{sec:ApplicationsOfXai} demonstrates the large and increasing number of applications of \ac{XAI} methods, techniques, and tools and their utility in real-world scenarios.

\subsection{XAI Trends, Advances, and Breakthroughs}
\label{sec:breakthroughs}
The prime goal of explanations is to make a model understandable or comprehensible to its stakeholders \cite{Paez2019Pragmatic, Langer2021What, Koehl2019Explainability, Chazette2021Exploring}. 
To this end, several methods have been introduced in the last few years to explain the decisions of complex \ac{AI} systems in many application domains \cite{Speith2022Review, Langer2021What, adadi2018peeking, Guidotti2018, carvalho2019mac}. 
Synthesizing explanations for \ac{AI} systems has been shown to have the potential to solve several technical and societal problems. 
Explanations can facilitate the understanding of how learning from data has occurred, for instance, via feature attribution methods. 
Furthermore, explanations can reveal information about how a model can be exploited to improve its performance. 
They can also support and improve human confidence in the output of a given model. 
Explanations may reveal the existence of hidden biases in the training data, learned during model training, that negatively impact a model's generalisation when predicting unseen data \cite{rawal2021recent}. 
Other purposes for demanding explanations include data stream settings, where they can be used to characterize what a model observes over time. This can serve as a knowledge base to detect non-stationarities in the task being solved and, thus, concept deviations \cite{hinder2020counterfactual}. 
Similarly, wrongly annotated data instances in large-scale databases can be identified by computing a measure of disagreement between the explanations issued for a model. 
Application opportunities such as these may also arise in vertical federated learning, where aggregation policies can be adjusted by examining commonalities among local models during update rounds \cite{khan2022vertical}. 
Explanations can also drive pruning and model compression strategies, linking irrelevant concepts to specific neurons that can hence be removed from a neural network \cite{yeom2021pruning}.

\subsubsection{Attribution Methods}
A lot of work exists on explaining the decisions of a classifier with \textit{attribution methods }\cite{samek2021explaining}. 
For instance, model agnostic attribution methods such as \ac{LIME} \cite{ribeiro_why_2016}, \ac{SHAP} \cite{lundberg_unified_2017}, and many others can contribute to the interpretation of \ac{DL} models by computing the importance of input features \cite{garreau2020explaining, slack2020fooling}. 
Furthermore, saliency maps built by attribution methods such as network gradients, Deconvolutional Neural Networks (DeConvNet), \ac{LRP}, Pattern Attribution, and Randomized Input Sampling for Explanation (RISE) can identify relevant inputs for the decisions of classification or regression tasks. 
In the image or text domain, explanations using attributions are intuitive and often perceived as easy to understand by the human receiver. 
For instance, one immediately understands that a classifier might not work correctly if it classifies horse images not by looking at the horse itself but by focusing on a copyright watermark, which is often present in images of this category. 
Such misbehaving classifiers have been termed \enquote*{Clever Hans} predictors \cite{lapuschkin2019unmasking} or \enquote*{Short-Cuts} \cite{Geirhos2020}.  
However, identifying such misbehaviour or understanding the meaning of an attribution-based explanation can be significantly more difficult in other domains \cite{Speith2022How}. 
For instance, an attribution map computed on a multivariate time series signal or a complex biological sequence can be significantly more difficult to understand for the human receiver; that means the \enquote*{interpretation gap} is much larger than in the horse example. 
Moreover, even in the image domain, attribution maps only indicate where the relevant information is located, but it is still up to the human to assign meaning to this information. 
For example, when an attribution map highlights the teeth of a 20 years old person as an indicator for the prediction of the class \enquote*{young adult}, it does not convey whether the white colour of the teeth is the important cue for the prediction or the fact that the person smiles \cite{lapuschkin2017understanding}.

\subsubsection{Interpretable Models}

Explainability in contexts like finance often has a special flavour. In this domain, information is mainly presented as tabular or temporal data. Here, traditional \ac{ML} techniques are often adopted, especially techniques based on \acp{DT} \cite{grinsztajn2022tree}. The benefit of these techniques is, among others, that supposedly they lead to inherently interpretable models. 
Some scholars argue that using a black box model, usually derived by applying \ac{DL} methods, only marginally improves the performance of classical \ac{AI} methods \cite{rudin2019stop} (cf., \cite{Crook2023Revisiting}). 
Accordingly, models that are interpretable by design, such as \acp{DT} \cite{rokach2016decision}, are preferred for many applications \cite{hatwell2020chirps}. 
For this reason, another recent development within \ac{XAI} is that of rule-based approaches and rule extraction methods, building on their long history within \ac{AI}. 
For example, using symbolic rules to derive knowledge is still popular today \cite{furnkranz2020cognitive}.
Although these methods can improve the overall performance of \ac{XAI} systems by synthesizing effective explanations, they are still largely ignored when prioritizing interpretability. 
One reason might be that the coverage and specificity of the generated trees or rules are low. 
In methods based on rule extraction, an opaque \enquote*{black box} model is typically trained first and then used to construct a transparent \enquote*{white box} model, such as a rule-based model or a \ac{DT}. 
However, limiting the complexity of a \ac{DT} while achieving a high accuracy via rule extraction is an open problem \cite{Krakauer2023}.

\subsubsection{New Kinds of Approaches}
Recent approaches have shown potential for resolving problems of older approaches, even if more research must be performed to confirm this \cite{huang2020tabtransformer, gorishniy2021revisiting}. One such approach has integrated attention-based explanations into a neural architecture to achieve an efficient computation of tabular data and to increase its interpretability \cite{arik2021tabnet}. 
Results are encouraging, but explanations remain highly subject to the inner variability of attention when transformer architectures are used. 
In that respect, the attention mechanisms could be heavily exploited with a variety of established techniques, including attention flow and rollout \cite{abnar2020quantifying}, \ac{LRP} adaptation \cite{ali2022xai}, or attention memory \cite{deb2023atman}. 
Such techniques are promising in enhancing explanations for complex models 
but the properties of explanation need to be further investigated, especially concerning stability, robustness, and fidelity \cite{DBLP:journals/semweb/Lecue20, Speith2022How, Speith2023New}. 
Connected to the use of rules as a means for enabling the explainability of \ac{AI} systems, another new trend within \ac{XAI} is the use of argumentation \cite{vcyras2021argumentative, Baum2018From, Baum2018Towards}. 
In particular, computational argumentation can be useful to explain all the steps towards a rational decision, as well as enabling reasoning under uncertainty to find solutions with conflictual pieces of information \cite{vassiliades2021argumentation, longo2016argumentation, Zeng2018Building}. 
In this context, rules are seen as arguments, and their interaction is seen as a conflict that can be resolved with argumentation semantics \cite{baroni2011introduction}. 
Typically, computational argumentation implements non-monotonic reasoning, a type of reasoning where conclusions can be retracted in the light of new reasons \cite{RizzoL18Inferential, rizzo2018comparative, longo2021examining}. 
This formalism is appealing within \ac{XAI} because it mirrors one-way human reasoning works \cite{Baum2018Towards}.

\subsection{Applications of XAI Methods}
\label{sec:ApplicationsOfXai}

\ac{XAI} methods have been widely applied in several fields, including finance, education, environmental science and agriculture, and medicine and health care. 
This section describes some of the many applications of \ac{XAI} methods. The goal is to provide stakeholders with illustrations and case studies.

\subsubsection{Medicine, Health-Care, and Bioinformatics}
The inferences produced by \ac{AI}-based systems, such as Clinical Decision Support Systems, are often used by doctors and clinicians to inform decision-making, communicate diagnoses to patients, and choose treatment decisions.
However, it is essential to adequately trust an \ac{AI}-supported medical decision, as, for example, a wrong diagnosis can have a significant impact on patients. 
In this regard, understanding \ac{AI}-supported decisions can help to calibrate trust and reliance.
For this reason, many \ac{XAI} methods such as \ac{LIME}, \ac{SHAP}, and Anchors have been applied in Electronic Medical Records, COVID-19 identification, chronic kidney disease, and fungal or bloodstream infections \cite{SBAND2023101286}. 
In these high-stakes scenarios, there is evidence that \ac{AI}-based systems can have superior diagnostic capabilities than human experts \cite{tschandl2019comparison}.
Thus, the explainability of these systems is not only a technological issue, but boils down to medical, legal, ethical, and societal questions that need careful consideration \cite{amann2020explainability}.

\subsubsection{Finance}
In finance, institutions such as banks and investment firms leverage \ac{AI} to automate their processes, reduce costs, improve service security, and, generally, gain a competitive advantage. 
\ac{AI} algorithms are used at scale to predict credit risk, detect fraud, and diagnose investment portfolios for optimisation purposes.
In these contexts, applying \ac{AI} often requires transparency and explainability for legal reasons.
This requirement is particularly significant in the customer banking sector, where banks must comply with strict regulations such as the USA Equal Credit Opportunity Act (ECOA) or the USA Fair Housing Act (FHA) to expose adverse action codes and provide clear explanations for their decisions. 
Similar guidelines and law enforcement are present in Europe guided by the General Data Protection Regulation (GDPR) law of the European Union. 
For example, if a customer's loan application is denied, the bank must be able to provide a clear and understandable reason for this.
It becomes increasingly difficult for banks, when adopting \ac{AI} algorithms, to provide explanations which are stable and trustworthy \cite{han2022explanation,agarwal2022openxai} In other words, it becomes increasingly hard to justify the inferences of \ac{AI} models, both with simpler transparent models \cite{bussmann2021explainable, sachan2020explainable, rudin2019we}, and even more with complex models \cite{arik2021tabnet, huang2020tabtransformer, gorishniy2021revisiting}. 
This lack of transparency can put banks at risk of regulatory penalties and erode customer trust.
In investment banking, the demand for \ac{XAI} is driven by the need to ensure the robustness and stability of \ac{AI} systems  \cite{mishra2021survey}, which could be subjected to extreme market conditions and unexpected events. 
If an \ac{AI} system makes inferences that are difficult to validate, it could lead to disastrous outcomes. 

\subsubsection{Environmental Science and Agriculture}
Another area of application of \ac{AI} that has benefited from adopting \ac{XAI} methods is the intelligent analysis, modelling, and management of agricultural and forest ecosystems---an important task for securing our planet for future generations.
For example, forest carbon stock is a critical metric for climate research and management, as forests play a vital role in sequestering atmospheric carbon dioxide. In this context, drones can be deployed for data collection, and \ac{ML} techniques can be used for estimating forest carbon storage \cite{sharma2022drones}. 
Forest inventory also plays a crucial role in forest engineering, as it provides critical information on forest characteristics, such as tree species, size, and density, which can inform forest management decisions \cite{mollmann2017practical, gollob2020forest}. 
In these life-critical environments, sensor-based technology is employed to collect data, which is often high-dimensional and heterogeneous, and then \ac{AI}-based models are trained on it. 
However, data is often poor in quality, thus leading to models that lack robustness. 
Furthermore, even if such models are robust, there are still challenges in terms of tracing and understanding their inferences, and in ascertaining the causal factors that underlie them.
Even the smallest perturbations in the input data can dramatically affect a model's output, leading to completely different inferences and thus undermining the trustworthiness of such models \cite{holzinger2022digital, holzinger2022challenges}.
Additionally, in these naturalistic environments, a challenge for forest engineering is the development of methods for uncertainty quantification and propagation. 
In fact, \ac{AI} methods for developing forest inventory models are subject to various sources of uncertainty, including measurement error, spatial variability, and model misspecification. 
It is, therefore, extremely important to analyse the robustness of \ac{AI} methods---for instance, through explainability---and enhance it for the produced models and their inferences \cite{holzinger2022next, holzinger2022information}.

\subsubsection{Education}
\ac{AIED} focuses on developing \ac{AI}-powered educational technologies to aid students, instructors, and educational institutions \cite{luckin2016intelligence, zawacki2019systematic, Longo2019Empowering} in their teaching and learning activities. 
On the one hand, for students, \ac{AIED} has focused on developing models \cite{desmarais2012review}, and adaptive systems that can identify learners' strengths and weaknesses across a variety of topics, leading to customized instructions and resources that align with their learning needs \cite{vanlehn2011relative}. 
These are, for example, focused on improving their meta-cognitive processes of self-monitoring, reflection, and planning \cite{bull2020there}.
On the other hand, for instructors, \ac{AIED} tools can act as smart teaching assistants \cite{deboulay2016}, help them orchestrate the classroom \cite{holstein2019co}, grade assessments \cite{singh2017gradescope}, and answer student queries \cite{hiremath2018chatbot}, minimizing students dropout \cite{liz2019systematic}. 
The most recent example of an application of \ac{AIED} includes the use of some \ac{LLM} capable of generating new textual content based on human input prompts. 
This can be used for writing essays, producing software code, or generating educational content such as multiple-choice questions or worked examples with step-by-step solutions. 
A growing concern is that students, instructors, and educators lose control of \ac{AI}-based technologies as they fail to determine how these work, why they produce certain outputs, and what impact they may have.
In particular, \ac{AIED} tools such as educational recommender systems are increasingly used to automate and personalize learning activities \cite{khosravi2019ripple}. 
These tools pose various concerns about their use in high-stakes decisions, including fairness, accountability, transparency, and ethics \cite{holmes2021ethics, baker2021algorithmic, kizilcec2022algorithmic}. 
The impact of these technologies on students' agency and self-regulated learning is a growing concern, as the lack of transparency and feedback can make it difficult for instructors and learners to calibrate their trust in \ac{AI}-based inferential systems and understand their current state of learning and the benefits derived from engaging with a particular educational resource \cite{abdi2020complementing}.

\section{Challenges and Research Directions}
\label{sec:challengesResearchDirections}

\begin{figure}[!ht]
    \centering
    \includegraphics[scale=0.52]{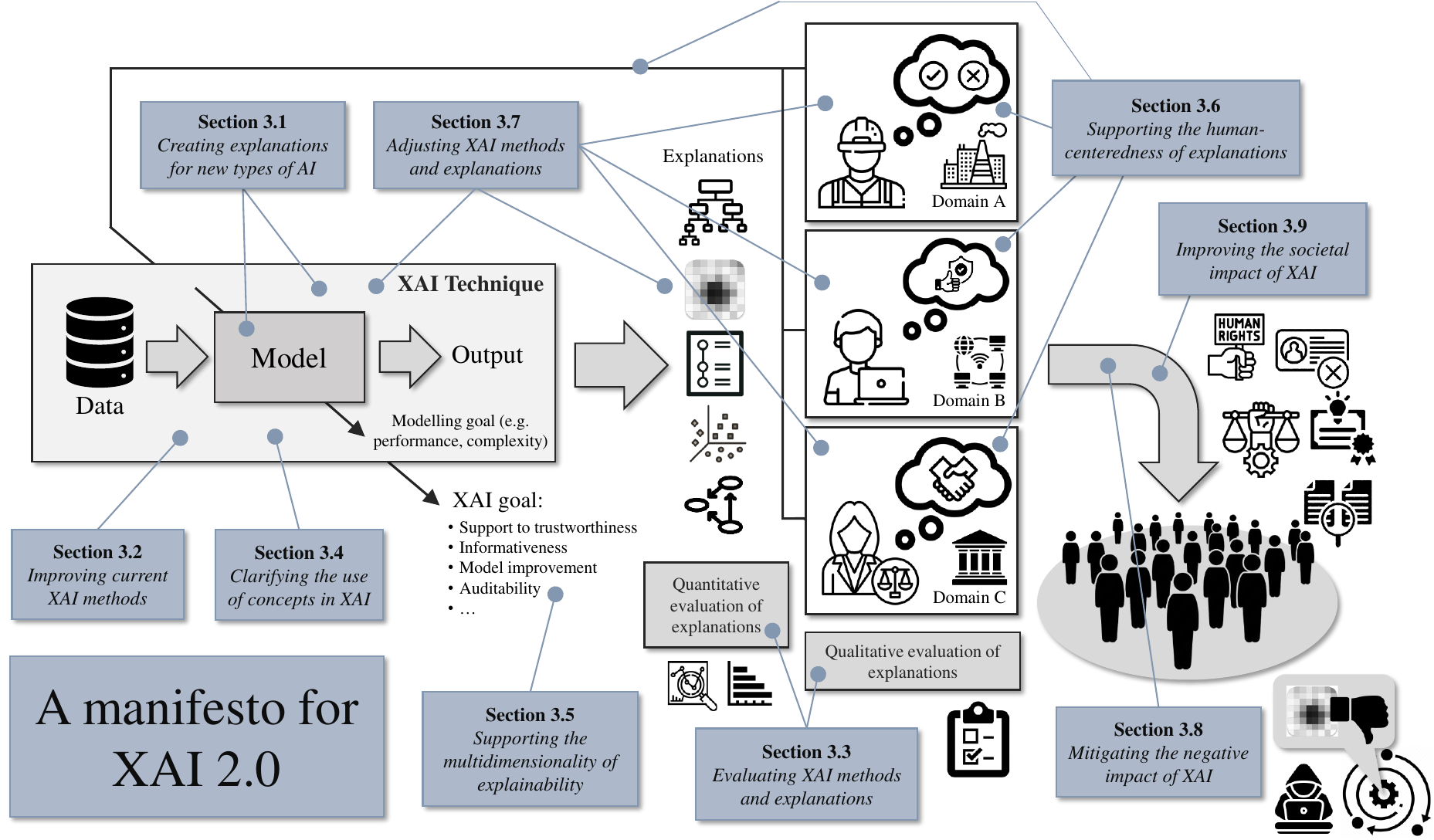}
    \caption{A manifesto for eXplainable Artificial Intelligence (XAI): High-level challenges}
    \label{fig:xAIChallenges}
\end{figure}

Despite the many advances, breakthroughs, and potential applications of \ac{XAI} methods, more research is clearly required to address open problems in the field. 
For example, it is still unclear how \ac{XAI} methods should be evaluated, how different terms should be used in the debate, or how, exactly, \ac{XAI} is related to trustworthiness. 
Many surveys tackling some of these aspects of \ac{XAI} exist and keep appearing in conference proceedings and journals \cite{ali2023explainable}. 
However, they are rather scattered, often specific to an application domain or focused on specific methods. 
Against this backdrop, this section aims at extracting and synthesizing the diverse challenges in \ac{XAI} that motivate the formation of a manifesto. 
Overall, we identified 27 problems, which we have grouped into nine high-level categories---our manifesto---as depicted in \ref{fig:xAIChallenges}. These problems and the related challenges are often interconnected; thus, they may, in principle, belong to multiple categories.

\subsection{Creating Explanations for New Types of AI}
The ever-evolving landscape of \ac{AI} introduces novel types of models, such as generative models or concept-based learning algorithms, each with its unique set of properties. Against this backdrop, this category of challenges describes the intricacies of creating explanations for these new types of \acp{AI}.

\subsubsection{Creating Explanations for Generative and Large Language Models}

Generative \ac{AI} models, such as those employed for diffusion denoising \cite{cro22, yang2022diffusion} or the family of GPT models for large-scale language generation \cite{topal2021exploring}, are disrupting many sectors. 
These models deliver exceptional performance due to their immense scale.
With billions, and in some cases, nearly trillions of parameters, however, their sheer size poses a significant challenge to existing \ac{XAI} methods \cite{zini2022explainability}. 
In particular, these methods grapple with the high-dimensional nature of such models, both in terms of computational complexity and in extracting learned concepts. 
For instance, one obstacle related to the latter point lies in the polysemantic nature of the neurons in generative models, which is thought to arise from a superposition of multiple independent features \cite{bricken_towards_2023}.
\ac{XAI} methods have so far been mostly limited to classification and regression problems. Accordingly, completely new approaches have to be developed for generative models. 
In particular, self-supervised or neural generative models such as Variational Autoencoders (VAEs) and Generative Adversarial Networks (GANs) are becoming more popular. For instance, examining the latent spaces they learn and synthesizing explanations for them is very challenging.
Another challenge, particularly for \acp{LLM}, concerns scaling laws. 
Neural scaling laws are functional relationships that relate variables associated to a neural network, such as the number of layers in its architecture and its achieved accuracy after  training. Such functional relation of two $x$ and $y$ variable is of the general form $y=a\cdot x^{\alpha}$, where $a$ and $\alpha$ are constants of the scaling law.
Such laws govern the aggregate capabilities of \acp{LLM}, yet a precise understanding of individual task-level implications of these laws remains elusive, as they appear to manifest unpredictably. It is an open issue whether scaling laws can be used to infer the quality of the artefacts or concepts learned by \acp{LLM}. Even if this were possible, such laws might reveal plateaus of \ac{XAI} scaling relative to general ability scaling or indicate a gap between the two.

\paragraph*{Solution Ideas}
Mechanistic interpretability \cite{Cammarata2020Thread, elhage_mathematical_2021} is a promising approach to gain deeper insights into the functioning and scaling laws of generative models, such as for grokking mechanics \cite{nanda_progress_2023} and the ability to solve problems recursively \cite{zhang_can_2023}. In particular, mechanistic interpretability has shown promising results at small model scales and for toy problems.
Researchers from institutions and companies like MIT, OpenAI, DeepMind, and Anthropic pursue mechanistic interpretability as an approach that attempts to reverse engineer the learned representations and algorithms of trained models using causality-based methods. 
Piecewise linear activation functions have been used to partition the activation space into polytope-shaped monosemantic regions \cite{black_interpreting_2022} and sparse autoencoders have been successfully used for the mono-semanticity of \ac{DNN} models \cite{bricken_towards_2023}.
There are also challenges in mechanistic interpretability, such as disentangling multiple algorithm implementations and finding unknown algorithms \cite{zhong_clock_2023}. Furthermore, there are preliminary results from vision models that scaling does not help the mechanistic interpretability of models \cite{zimmermann_scale_2023}, calling for designing models for mechanistic interpretability. 
A potential complement to mechanistic interpretability may be information geometry \cite{amari_information_2016}, which can help analyze high-dimensional spaces involved in the processing of \acp{LLM}. Furthermore, constraints may have to be imposed on the training and functioning of \acp{LLM} to ensure safety as well as explainability \cite{brcic_impossibility_2023, zimmermann_scale_2023}. Such constraints could be directly part of the automated optimization (learning) process or indirectly used through a human-in-the-loop approach. An example of good direction is in \cite{liu_seeing_2023} which introduces a training procedure that encourages modularity and interpretability by discouraging non-local connections between neurons through local L1 regularization with swaps of neuron locations. Finally, it remains to be seen if these methods can be scaled to relevant models and problem sizes and complexities.

\subsubsection{Creating Explanations for Concept-Based Learning Algorithms}
Concept-based learning algorithms are another class of new forms of \ac{AI} for which good \ac{XAI} methods do not yet exist. Several such algorithms have been proposed over the years to directly learn features that describe \enquote*{prototypical concepts} or \enquote*{prototypes} present in each input to the model, including ProtoPNet \cite{chen2019looks}, ProtoTree \cite{nauta2021neural}, ProtoPShare \cite{rymarczyk2021protopshare}, Concept Bottleneck Models \cite{koh2020concept}, Concept Activation Vectors \cite{kim2018interpretability}, Concept Embedding Models \cite{zarlenga2022concept} or Concept Atlases \cite{achtibat_attribution_2023}. 
Neuro-symbolic learning, namely, the symbiosis between connectionist and concept-based symbolic learning, has recently also gained momentum \cite{mao2019neuro, sarker2021neuro, hamilton2022neuro}. Hybridizing knowledge graphs (KGs) with learning algorithms also fall within the landscape of approaches used to map knowledge encoded in the parameters of a model with a priori known concepts and the interrelationships among them \cite{tiddi2022knowledge}.
Unfortunately, use cases proposed to showcase how these approaches explain their decisions are limited, very narrow, and assume a priori knowledge about the concepts that can be discriminative for the task at hand. This assumption may imprint a large inductive bias in their explanation-producing process, not properly generalising when explaining new inputs that are distributionally novel with respect to the training data. Furthermore, the continuous proposal of new datasets for concept learning, including Clevr/Clevrer \cite{johnson2017clevr,yi2019clevrer}, Kandinsky Patterns \cite{muller2021kandinsky} or Closure \cite{bahdanau2019closure}, sheds evidence on the need for eliciting local explanations that can be formulated in terms of concepts and their spatial distribution. 

\paragraph*{Solution Ideas}
One interesting avenue of research explores the potential for genetically evolvable connections between identifiable concepts in input data using object detection models and evolutionary programming solvers \cite{mei2022explainable}. This hybridization could offer the advantage of employing symbolic classifiers that are interpretable, algorithmically transparent, and well-suited for handling datasets that encapsulate discriminative, concept-wise compositional information. Additionally, there is a growing demand to expand hybrid approaches that unite Knowledge Graphs with concept-based learning methods. This expansion aims to enable the discovery of relevant concepts, attributes, and relationships that extend beyond the confines of specific use cases or domains, as discussed by Lecue et al. \cite{DBLP:journals/semweb/Lecue20}.

\subsection{Improving Current XAI Methods}
A spectrum of challenges arises when considering current \ac{XAI} methods. Many of these have long-known disadvantages that need to be overcome, as described below.

\subsubsection{Augmenting and Improving Attribution Methods}
One major branch of \ac{XAI} methods relies on pixel attribution with heatmaps or saliency masks \cite{sim13}, one of the most prominent classes of \ac{XAI} methods used for computer vision tasks. Such methods are often based on perturbations \cite{ribeiro2016should, zeil14} or gradients \cite{selvaraju2017grad, bach2015pixel}. Despite the great success of these methods to, for instance, detect biases and flaws in the learned prediction strategies (so-called \enquote*{Clever Hans Effect} \cite{lapuschkin2019unmasking}, see \ref{sec:breakthroughs}), attribution-based explanation methods also have limitations. For instance, saliency masks on the level of pixels are often unsuited for laypersons \cite{Speith2022How}. 
A major technical limitation of attribution methods is their sensitivity to 1) internal hyper-parameter tuning and customisation (such as baselines), 2) the manual settings of interpretable interfaces, and 3) the assumption surrounding the model under exploitation. For example, the results of model-agnostic attribution methods, including \ac{LIME} and \ac{SHAP}, can change based on the range of input perturbation. Similarly, many gradient-based methods require setting a proper sampling interval. Finally, relevance propagation methods such as \ac{LRP} have to use different methods depending on the layer of the \ac{DNN}. Additionally, some methods have issues with computational efficiency, requiring many passes for calculating attributions.

\paragraph*{Solution Ideas}
We can tackle the inherent issues in attribution methods by combining them with other approaches to XAI to get a portfolio approach that hedges the weak characteristics of each individual approach. Mechanistic interpretability is an orthogonal approach with different characteristics that could play along well with these approaches. Methods in the portfolio could negotiate like in a market to tune themselves and converge to a majority view, or even better, a list of hypotheses with their plausibilities based on the votes of each portfolio participant.

\subsubsection{Removing Artifacts in Synthesis-Based Explanations}
Generating explanations through synthesis is a promising direction to advance the field of \ac{XAI}. While a user is unlikely to directly understand the layer activations of specific classes, it may be different for examples of those classes.
The synthesis of such examples, however, is often noisy. For instance, a synthesized image may contain artefacts. It is unclear whether this noise is due to the synthesis process itself or is, de facto, part of a concept learned by the model.
For example, while a GAN architecture can synthesize an image representing the pattern that activates a neuron most strongly, this image might have various artefacts that make it appear somewhat distorted. 
This might happen due to shortcomings of the GAN models, which means these artefacts must actually be present to activate the neuron strongly. 
Two existing methods for synthesis in the literature are a decoder for layer activation \cite{schn22concept}, and a GAN for single neurons \cite{ngu16}. 
Unfortunately, the mere synthesis of inputs is insufficient for understanding concepts. There are few works on using generative models for explanations, including the work of \cite{schn22concept} and the chapter concept vectors in \cite{rau22}. Most methods explaining concepts rely on a given dataset of human-defined concepts \cite{rau22}, which, however, might not be available for a specific domain and must be collected at high costs. Furthermore, even if a dataset is available, there is a considerable risk that the user-defined concepts are incomplete or inaccurate, leading to poor or biased explanations.

\paragraph*{Solution Ideas}
To minimize artefacts, state-of-the-art models and recent popular techniques in \ac{DL} \cite{sch23survey}, especially diffusion models, could be leveraged \cite{cro22}. However, even state-of-the-art generative models do not ensure the absence of artefacts. 
Thus, to verify whether there are any distortions due to the synthesis, one idea is to compute a reconstruction of the original input serving as a reference \cite{sch22exp}. This reference stems from a separate model with the same architecture as the decoder synthesizing inputs from the model to explain. Subsequently, a user can compare the original input, the synthesized image from layer activations of the model to explain (that means, what the classifier \enquote*{sees}), and the reference, allowing them to identify distortions due to the synthesis process. If it can be seen that the original image and the reference are fairly similar, then distortions might be considered minor. However, a classifier might not rely on certain concepts associated with the input. Therefore, while the comparison with a reference might be considered a valid approach, it is still tedious for the lay user and non-trivial to apply beyond autoencoders.

\subsubsection{Creating Robust Explanations}
The fragility of posthoc \ac{XAI} methods to small perturbations at the model's input and the known inconsistency in synthesized explanations for a given input \cite{yeh2019fidelity} highlights the challenge of creating robust explanations. This is frequently advocated as a requirement for calibrating human trust and building acceptance of a model being audited. 
Several works have advocated the idea of exploiting explanations beyond just explaining decisions \cite{weber2022beyond, gao2022going}, for instance, to also improve models. However, the susceptibility of explanations to the \ac{XAI} technique under consideration detracts from the explanation's robustness, jeopardizing the reliable application of explanations to improve a model. 
Methodologies for delivering robust explanations under different circumstances are investigated in several recent works \cite{mishra2021survey, ferrario2022robustness, qiu2022generating}. A satisfying solution, however, does not yet exist. The difficulty lies especially in the fact that for a robust explanation, the model itself must be robust. 

\paragraph*{Solution Ideas}
As a first step towards robust explanations, evaluations on standard benchmarks should be done to identify common biases of an \ac{XAI} method and to define ways to mitigate them. Furthermore, robust explanations could be created by aggregating explanations. For example, a proposal exists to blend uncertainty quantification and \ac{XAI} \cite{seuss2021bridging}. 
Other research has placed emphasis on the robustness of the \ac{AI} model itself, for instance, in the form of explanations that inform about model inversion or extraction attacks \cite{kuppa2020black, oksuz2023autolycus}. In a similar vein, the recently proposed \enquote{reveal to revise} framework enables practitioners to iteratively identify, mitigate, and (re-)evaluate spurious model behaviour with a minimal amount of human interaction \cite{PahArXiv23b}.

\subsection{Evaluating XAI Methods and Explanations}
Evaluation is an important aspect of the development and deployment of \ac{XAI} systems. Evaluating \ac{XAI} methods, however, is a complex task, and no gold standard exists on what makes for a good explanation \cite{Speith2022How}.

\subsubsection{Facilitating Human Evaluation of Explanations}
One problem concerning the evaluation of \ac{XAI} methods is that they often lack user studies.
Current evaluation approaches typically only analyze certain properties of the \ac{XAI} methods themselves without accounting for the interaction with the final user \cite{abs-2102-13076, Guidotti21, guidotti2022counterfactual, Vilone2021Notions, Speith2023New}. For instance, a survey of user studies has shown that only 36 out of 127 research works employing counterfactual explainers adopted a human evaluation approach, and only 7\% of them tested alternative approaches \cite{keane2021if}.
Individual differences in understanding, prior knowledge, and the cognitive load required to comprehend explanations add further challenges to evaluating \ac{XAI} methods. In general, it is difficult to compare different forms and types of explanations to determine which of them are the most effective. Additionally, users are typically \enquote*{passive recipients} of explanations, and the actual usage or exploitation of such explanations is barely tested. For certain properties, there are no approaches at all that test for them \cite{Speith2023New}. While some studies evaluated the impact of synthesized explanations of \ac{AI} systems on humans when compared to the scenario where no explanations were provided \cite{DBLP:conf/iui/DodgeLZBD19, DBLP:conf/fat/LucicHR20,DBLP:conf/hhai/MettaGYGR22}, there is clearly a need for more (and more systematic) work on the topic.

\paragraph*{Solution Ideas}
Establishing a solid foundation for \ac{XAI} must be grounded in empirical research involving users. Achieving this demands a collaborative, interdisciplinary approach, uniting \ac{ML} experts with researchers from fields like HCI, psychology, and the social sciences. Valuable insights can be gleaned from the collective body of knowledge in these domains, leveraging their expertise in conducting user studies \cite{Miller2017Explainable, Miller2019Explanation}. 
To streamline the evaluation process, it is imperative to establish standardized frameworks encompassing every stage of it, from formulating hypotheses to data collection, analysis, and utilising online questionnaires. With this robust methodology in place, the research community can, then, embark on the crucial task of developing heuristics, principles, and patterns that enable the design of effective \ac{XAI} systems for real-world applications. This comprehensive approach ensures that \ac{XAI} not only benefits from theoretical foundations but is also shaped by empirical user-centric research, ultimately enhancing its practical utility.

\subsubsection{Creating an Evaluation Framework for XAI Methods}
There are several works that address the evaluation of \ac{XAI} methods. Hoffman et al.~\cite{Hoffman2018MetricsFE}, for instance, integrate extensive literature and various psychometric assessments to introduce key concepts for measuring the quality of an \ac{XAI} system. Similarly, Vilone and Longo \cite{VILONE202189} aggregate evaluation approaches for \ac{XAI} methods from several scientific studies via a hierarchical system. Furthermore, Van der Lee et al.~\cite{VANDERLEE2021101151} define a list of steps and best practices for conducting evaluations in the context of generated text. 
An analysis of these works reveals that evaluating the goodness and effectiveness of explanations is a prerequisite for calibrating trust in \ac{AI}. However, 
there is currently a lack of standardized methods and metrics for evaluating \ac{XAI} systems. In other words, despite the broad interest in the design of \ac{XAI} methods \cite{GuidottiMRTGP19, abs-2102-13076, AdadiB18, carvalho2019mac, ArrietaRSBTBGGM20}, 
it is still unclear how to compare the results of different evaluations and establish a common understanding of how to evaluate explanations. 
What is missing is a set of evaluation metrics for explainability that are generally applicable across studies, contexts, and settings.

\paragraph*{Solution Ideas}
There are already some good approaches in the literature to solve this problem. In a recent survey on the evaluation of \ac{XAI}, for instance, the authors identify several conceptual properties that should be considered to assess the quality of an explanation, and they propose quantitative evaluation methods to evaluate an explanation \cite{EvalXAISurvey2023}. Furthermore, the recently developed \ac{XAI} evaluation framework Quantus \cite{HedJMLR23} implements over 30 evaluation metrics from six categories. 
Frameworks such as Quantus allow for the evaluation and comparison of explanations in a standardized and reproducible manner. Furthermore, publicly available \ac{XAI} evaluation datasets with ground truth information, such as CLEVR-XAI \cite{ArrINF22}, allow for objective evaluations. In the future, artefacts like these need to be extended to more application areas, especially outside of computer vision.

\subsubsection{Overcoming Limitations of Studies with Humans}
Evaluating \ac{XAI} methods with humans has limitations. Often, the number of participants that can be put together in a study does not represent the general population. Thus, a study's results may be prone to bias and errors and may not generalise well \cite{Frederik2023}. Overall, the evaluation of \ac{XAI} methods in studies with humans is prone to issues such as poor reproducibility and inappropriate statistical analyses, resulting in no solid evidence for their usefulness \cite{DBLP:conf/iui/DodgeLZBD19, DBLP:conf/fat/LucicHR20, DBLP:conf/hhai/MettaGYGR22, longo2012formalising, longo2015designing, hancock2021mental, Longo2022}.

\paragraph*{Solution Ideas}
A potential solution involves augmenting human studies with synthetic data and virtual participants. By creating synthetic datasets that span a wide range of demographic characteristics, behaviours, and preferences, researchers can address the issue of limited sample representativeness. These synthetic datasets can simulate diverse user profiles and scenarios, enabling more robust and extensive evaluations of \ac{XAI} methods. Additionally, virtual participants, based on \ac{AI}-driven agents or personas, can be incorporated into studies to provide a broader range of user interactions and perspectives.
To enhance the reproducibility and rigour of \ac{XAI} evaluations, standardized methodologies and statistical analyses must be employed. Researchers should adopt transparent reporting practices and adhere to well-defined evaluation protocols, ensuring that the evidence generated from these studies is solid and dependable. Another approach would be to create sample explanations or schemes for explanations against which generated explanations are checked. While the samples would still need to be tested in studies first, this could alleviate the overall need for studies with humans.

\subsection{Clarifying the Use of Concepts in XAI}
As a multidisciplinary research area, another category of challenges for \ac{XAI} is the disparate and unclear use of terms.

\subsubsection{Elucidating the Main Concepts}
\label{sec:conceptual}

In research on \ac{XAI}, there is a conceptual ambiguity regarding various terms, such as explainability, interpretability, transparency, understanding, explicability, perspicuity, and intelligibility. This represents a challenge in \ac{XAI}, as the lack of clear and consistent definitions of terms can hinder progress in developing effective and useful \ac{XAI} systems.
Some researchers have attempted to define an explainable Artificial System such as one that produces details or reasons to make its functioning clear or easy to understand \cite{Barredo2020Explainable}.
Other researchers use terms like explainability and interpretability synonymously \cite{Miller2019Explanation, Arya2019Explanation}, while others draw major distinctions between them \cite{rudin2019stop}. These differences pose problems for applied research and interdisciplinary collaboration.
Discussions about clarifying terms in the field of \ac{XAI} tend to take two distinct approaches. On the one hand, some contend that attempts to define the terms in question are futile, impossible, counterproductive or unnecessary, and previous definitions of explainability have failed and, in general, the whole endeavour of finding definitions is doomed to failure (for example, \cite{Krishnan2020Against, lipton2018myth, Ehsan2022Social}). On the other hand, some attempt to provide explicit definitions, intending to differentiate between the various terms employed (for example, \cite{Clinciu2019Survey, Graziani2022Global, Vilone2021Notions, Chakraborti2019Explicability, Sterz2021Towards}).

\paragraph*{Solution Ideas}
As the lack of a clear and consistent definition of terms related to explainability can hinder progress in developing effective and useful \ac{XAI} systems, the communication challenges should be addressed holistically rather than perpetuated by ambiguity in the use of terms. Against this background, it seems desirable to join the latter of the above camps and strive for a uniform use of different terms. A minimal solution of this kind would be for authors to always clarify, in their articles, what they mean by certain concepts. 
A more desirable solution, however, would be to define the various terms once and for all. In this line of thought, meaningful definitions can only be found if already existing ways of usage are considered. Creating completely new usages of the various terms is more likely to contribute to conceptual confusion than to resolve it. The first step in coining a generally applicable definition of the terms is, therefore, to identify current usages of them and to create an overview and comparison of them. For instance, some work identifies relevant notions \cite{Vilone2021Notions}, but limited work exists in comparing them. As a next step, the merit of the various proposed definitions must be determined. For this purpose, quality criteria should be established (see, for instance, \cite{Clinciu2019Survey}), which can be consulted to evaluate each proposed definition.

\subsubsection{Clarifying the Relationship Between XAI and Trustworthiness}

A similar conceptual challenge exists concerning trustworthiness. Properties like safety, fairness, and accountability are often mentioned for meeting regulatory actions focusing on the trustworthiness of \ac{AI}. For instance, the Ethics Guidelines for Trustworthy \ac{AI}, issued by the EU High-Level Expert Group on \ac{AI}, listed seven requirements for \ac{AI}-based models and systems to be seen as trustworthy \cite{AI-EU:19}: human agency and oversight, technical robustness and safety, privacy awareness and data governance, transparency and explainability, diversity, non-discrimination and fairness, societal and environmental well-being, and accountability. While \ac{XAI} has the potential to help with most of these \cite{Kaestner2021Relation}, it is taken to help with one of them primarily: transparency and explainability. 
However, even this relationship is unclear, as many sources contain contradicting statements. In these sources, it is possible to observe various claims about the relationship between trustworthiness and \ac{XAI}: trustworthiness is seen as a main goal of \ac{XAI} \cite{Barredo2020Explainable}, but \ac{XAI} is also claimed to be a part of trustworthiness \cite{Robbins2019Misdirected}. \ac{XAI} is purported to change the belief in the trustworthiness of a system \cite{Kizilcec2016How}, while it should also support the trustworthy integration of systems \cite{Ghosh2019Interpretable}. 
These are just a few examples, and in other articles, it is also possible to find completely different relationships \cite{Kaestner2021Relation, Ribeiro2016Why, Vilone2021Notions}. One reason for this divergence in descriptions is that there is no uniform way of using terms like trustworthiness (and other terms in \ac{XAI}, see \ref{sec:conceptual}). For this reason, as long as it is not clarified what each term describes and what property it expresses, it will not be possible to specify the relationship between \ac{XAI} and trustworthiness. 

\paragraph*{Solution Ideas}

The relationship between XAI and trustworthiness is widely discussed. We must distinguish between trustworthiness as a property of an AI system, trustworthy AI as a technology enabler to accomplish a responsible and safe AI, and the technical requirements required for an AI system to be trustworthy. XAI is identified as one of the seven trustworthy AI requirements \cite{AI-EU:19,DiazRodriguez2023}. On the other hand, XAI must contribute towards achieving trustworthiness. Currently, it is necessary to connect XAI with the fundamental properties of AI trustworthiness for AI risk management and the AI lifecycle, measuring their presence and impact. In this regard, we must highlight the report recently published by the UC Berkeley Center for Long-Term Cybersecurity (CLTC) \footnote{\url{https://cltc.berkeley.edu/publication/a-taxonomy-of-trustworthiness-for-artificial-intelligence/} (Last access: 2023/10/28)}. This report aims to help organizations develop and deploy more trustworthy AI technologies, including 150 properties related to one of the seven ``characteristics of trustworthiness'' defined in the NIST AI RMF\footnote{l{https://www.nist.gov/itl/ai-risk-management-framework} (Last access: 2023/10/28)}: valid and reliable, safe, secure and resilient, accountable and transparent, explainable and interpretable, privacy-enhanced, and fair with harmful biases managed.
Another important aspect is AI governance \cite{PALLADINO2023} and the need of governance measures linked to the importance of managing AI risks, another scenario where XAI becomes of utmost importance. These are new fundamental scenarios posing essential challenges for the design, development, and safe deployment of responsible AI systems \cite{DiazRodriguez2023}. In the current debate, XAI is identified as a vital technology to decrease the uncertainty and worry about AI systems in the society.

\subsubsection{Finding a Useful Account of Understanding}
Another challenge to bring about conceptual clarity is finding a useful account of understanding. An obstacle to providing such an account of understanding in \ac{XAI} is the lack of conceptual clarity about what understanding itself is. 
In philosophy, there are at least three different approaches to this problem. The more traditional view asserts that understanding logically depends on explanation: only \textit{true} explanations can provide understanding \cite{khalifa2012inaugurating, strevens2013no}. The other end of the spectrum is occupied by philosophers who allow other paths to understanding, even if they offer distorted or false accounts of their targets \cite{lipton2009understanding, elgin2017true}. 
Finally, intermediate views exist, allowing that some, but not all of the pieces of information used to provide understanding can be false \cite{kvanvig2009responses, mizrahi2012idealizations, carter2016objectual}. 
There is no consensus regarding which of these views is more adequate in the context of \ac{AI} explanations. For example, while \cite{erasmus2021interpretability} sides with the traditional view, \cite{Paez2019Pragmatic} adopts a more pragmatic stance.
Another obstacle arises from the fact that the understanding provided by \ac{XAI} methods that account for singular predictions need not be the same type of understanding provided by proxy or surrogate models that offer a global account of the target \ac{AI} model. There might be different underlying cognitive processes and abilities involved in each case. Prima facie, the explanation of singular predictions provides a type of understanding that epistemologists call \enquote*{understanding-why} \cite{pritchard2008knowing}, while proxy or surrogate models provide \enquote*{objectual explanations} \cite{zagzebski2009epistemology} of their targets.
The relation between the two types of understanding requires clarification, not only from a philosophical perspective but also from the point of view of psychology. In addition, there is a third type of understanding that depends entirely on the functional correlations between inputs and outputs \cite{lombrozo2019mechanistic}. Functional understanding might be sufficient for most users in many cases of human-computer interaction.

\paragraph*{Solution Ideas}
Solving the problem of a useful account of understanding in \ac{XAI} potentially requires a two-pronged approach. On the one hand, conceptual clarity is required. Several recent papers \cite{Paez2019Pragmatic, sullivan2022understanding, creel2020transparency, duran2021dissecting, erasmus2021interpretability, zednik2021solving, fleisher2022under, pirozelli2022sources, Chazette2021Exploring, Koehl2019Explainability} have focused on the relation between explanation and understanding in \ac{AI}. The conceptual map of this specific problem is now quite clear. Still, future developments will have to respond to new psychological evidence about human-computer interaction and to the development of new \ac{XAI} methods. 
On the other hand, empirical work on understanding is essential. For a long time, \ac{XAI} researchers have tried to ensure that the methods they develop are comprehensible to their peers, a phenomenon referred to as \enquote{the inmates running the asylum} \cite[p.~36]{Miller2017Explainable}. The proposed and endorsed alternative is to incorporate results from psychology and philosophy to \ac{XAI} \cite{Langer2021What, de2017people, Miller2019Explanation, mittelstadt2019explaining}. Existing theories of how people formulate questions and how they select and evaluate answers should inform the discussion \cite{Miller2019Explanation}.

\subsection{Supporting the Multi-Dimensionality of Explainability}
Another class of challenges for \ac{XAI} is that explanations are multi-dimensional. In other words, explainability is a concept which has multiple facets and spans a variety of disciplines.

\subsubsection{Creating Multi-Faceted Explanations}
For regulatory purposes, explanations should depend on and incorporate information about requirements for trustworthy \ac{AI} systems. In some cases, there is no reason to spend much resources and effort explaining a decision made by an \ac{AI} model if such a model is not accurate, not lawful, or not fair. 
In this line of thought, there have recently been calls stating that different dimensions of trustworthiness (for example, safety, fairness, accountability) should not be shown separately or individually to the audience of a given model or \ac{AI}-based artefact. For this reason, explanations should be offered to humans by not only \textit{explaining} the functioning (that means, traditional transparency and explainability) but also by \textit{justifying} the reliability of the inferences of an \ac{AI} system (for example, concerning technical robustness, safety, lawfulness, and fairness). If these properties are not considered, explanations will fail to calibrate users' trust correctly. This issue is particularly acute in situations of concept drifts or uncertainty. 

\paragraph*{Solution Ideas}
One approach to such multi-faceted explanations could involve developing trustworthiness metrics that encapsulate dimensions like safety, fairness, and accountability. \ac{XAI} can, then, be tailored to the trustworthiness level of the \ac{AI} system, ensuring that less trustworthy models provide extensive \textit{justifications} for their decisions while highly trustworthy systems may offer simple \textit{explanations}. Trustworthiness thresholds can be established, triggering detailed explanations when the system falls below a predefined trustworthiness level.
Furthermore, dynamic explanations that adapt to context, such as concept drift or uncertainty, can ensure that users' trust remains calibrated. A user-centric approach, allowing customization of explanation depth, would empower users to align the system's explanations with their specific needs. Transparency in the trustworthiness assessment process may enhance user confidence, and continuous monitoring and reporting offer the capability to adapt explanations as trustworthiness metrics change. This comprehensive strategy aims to ensure that trustworthiness considerations are an integral part of the \ac{XAI} process, leading to multi-faceted explanations.
A complementary way to tackle the multidimensionality of explanation concerns its operationalisation which should be performed as it happens with other psychological constructs such as \enquote*{intelligence} or \enquote*{cognitive load} \cite{Longo2022}. A solution is to propose a novel, inclusive definition of explainability that is modellable and that can be seen as a foundation to support the next generation of empirical-based research in the field. 
Modelability here means that the definition should contain high-level classes of notions and concepts that can be individually modelled, operationalized, and investigated empirically. The main rationale behind this solution is practical, as the aim is to provide scholars with an operational characterization of explainability that can be parsed into sub-components that, in turn, can be individually modelled. This should motivate the use of quantitative methods for greater reproducibility, replicability and falsifiability.

\subsubsection{Enabling Interdisciplinary Work in XAI}
\ac{XAI} is an interdisciplinary research field \cite{Langer2021What, Langer2021Auditing}. For example, through the collaboration of philosophers and computer scientists, \ac{XAI} is envisioned to ensure the ethical use of \ac{AI} \cite{Langer2021Auditing}. However, it is often difficult for researchers of different disciplines to engage in joint research in \ac{XAI} \cite{Speith2022Review}. 
There are several reasons for that. First, the rapid increase of publications in \ac{XAI} makes it difficult for researchers to keep up even with research in their own discipline, such that they often cannot spare to engage with research of other disciplines (which also has an overwhelming number of publications) \cite{Speith2022Review}. Furthermore, the different disciplines involved in \ac{XAI} may have their own established usage of certain terms \cite{Graziani2022Global}. This can lead to confusion and difficulty adapting to different usage in \ac{XAI}. 
Eventually, for terms for which there is no common usage, different disciplines may establish their own meanings, further leading to confusion.

\paragraph*{Solution Ideas}
To counteract the information overload caused by a rapid increase in publications, a centralized knowledge-sharing platform for \ac{XAI} could be established. This platform would curate and categorize relevant research from various disciplines, making it more manageable for scholars to access and engage with research from other disciplines. A crucial aspect of this collaborative platform would involve the development of standardized terminology and glossaries that unify the usage of key terms across disciplines. This would reduce confusion arising from varying interpretations of terminology, ensuring that researchers can communicate effectively and harmoniously. These terms should be updated periodically to accommodate evolving interdisciplinary insights.
Moreover, fostering regular cross-disciplinary dialogues and forums can promote mutual understanding among researchers from different backgrounds. Dedicated workshops, conferences, and seminars for interdisciplinary work in \ac{XAI} could facilitate knowledge exchange and encourage the development of shared research goals and methodologies.
Additionally, funding agencies and institutions should incentivize and prioritize interdisciplinary research by offering grants, awards, and recognition for collaborative projects. This would motivate researchers to actively engage in cross-disciplinary efforts in \ac{XAI}.

\subsection{Supporting the Human-Centeredness of Explanations}
One class of challenge in \ac{XAI} lies in providing explanations that are specifically adapted to the humans receiving them.

\subsubsection{Creating Human-Understandable Explanations}
In his seminal paper about explanations in \ac{AI} and social sciences, Miller points out that explanations should be social, contrastive, and selective to be understandable to humans \cite{Miller2019Explanation}. Confalonieri et al. discussed further properties for explanations, including integrating symbolic knowledge and statistical approaches to explainability \cite{WIREs-2020, AIJ-2021}. 
Unfortunately, many current \ac{XAI} methods do not have these properties. In particular, many \ac{XAI} methods provide explanations that do not extrapolate beyond the domain of their input data. A clear example of this phenomenon is the manifold number of gradient-based attribution methods \cite{nielsen2022robust, selvaraju2017grad, bach2015pixel}, all yielding explanations in the form of visual heatmaps quantifying the relative importance of every pixel of the input image to the prediction issued by the model. Many contributions assume that such heatmaps are enough for explainability simply because a \enquote*{narrative} can be built to relate pixels to concepts that emerge from intuition. 
There are, however, several problems with this assumption. First, the intuitions in question are often from experts \cite{Miller2017Explainable}, and the presentation of explanations in the form of pixel attributions may not be comprehensible to laypersons \cite{Speith2022How}.
Second, in more complex scenarios, crafting a narrative can become challenging, especially when discriminating between classes relies on intricate distributions of concepts within an image \cite{yuan2022compositional, klinger2020study}, or other semantically defined relations among the entities to which these concepts belong.
Third, these narratives are sometimes elaborate guesses at best. Assume a saliency map, serving as an explanation, highlights coarsely the face of a person to classify it as a human. It is unclear whether the underlying classifier used features such as the shape of the face, the skin colour of the face, or characteristics of the face such as mouth and lips, or a combination thereof, to make its inference.

\paragraph*{Solution Ideas}
Audiences without technical background are often concerned with \textit{concepts}, not with data. For instance, in a classifier discriminating between \enquote*{dogs} and \enquote*{cats}, it is significantly more informative for many people to state that \enquote*{the shape of whiskers} is a discriminative concept in the images rather than the relevance of isolated pixels as dictated by a gradient-based attribution technique. In this line of thought, concept-based \ac{XAI} methods explain individual predictions not as pixel-wise attributions but in terms of semantically meaningful concepts (for example, \enquote*{eye}, \enquote*{red stripe}, \enquote*{tyre}) represented by hidden-layer elements of the neural network. 
Often, concept-based explanations can be enriched by reference samples from the training dataset. Combining local \ac{XAI} methods (that means, explaining individual predictions) with global \ac{XAI} methods (that means, explaining the whole model) might lead to semantically richer and more human-understandable explanations. This \enquote*{glocal} approach was taken in concept relevance propagation, an upgrade to \ac{LRP}, to simultaneously identify concepts learned by the model (global) and match them to each individual input (local) \cite{achtibat_attribution_2023}.
Enriching explanations with explicit knowledge can enact scenarios in which formal and common-sense reasoning can be used to create explanations that are closer to the way in which humans think. In this line of thought, computational argumentation techniques could be exploited to generate explanations that can mimic the way humans reason under uncertainty \cite{rizzo2020empirical, longo2016argumentation, RizzoL18Explainability, vilone2022novel, ViloneLongo2022XAIArg, Baum2018From, Baum2018Towards, vassiliades2021argumentation}.
Another possible solution to create human-understandable explanations is to map explanations to a more interpretable domain. For instance, one approach to providing more interpretable explanations on time series data has been recently explored in \cite{vielhaben2023explainable}. In this context, the explanation is firstly computed on the time domain, which is the domain of the operation of the model. Then, the solution is mapped through an invertible layer where explanations can be computed in different spaces. Future research should investigate meaningful invertible mappings, for example, by using autoencoders \cite{infoLongo2023}, for this and other domains.

\subsubsection{Facilitating Explainability With Concept-Based Explanations}
Humans and \ac{AI} systems make decisions differently. In particular, \ac{AI} systems, especially those based on \ac{DL}, often rely on features that are hard to grasp for humans. On the other hand, humans use concepts that are coarse-grained representations of reality \cite{Quine1953-QUIOWT, krakauer_computational_2014}. This difference is often not taken into account when it comes to creating explanations. For example, prominent explainability methods such as \ac{LIME} or \ac{SHAP} rely on feature attributions that might reveal little about how an \ac{AI} model works \cite{garreau2020explaining, slack2020fooling}. 
Concept-based \ac{XAI} methods go beyond attribution and aim to express human-understandable concepts as part of the explanation that must first be synthesized from the model to be explained. One benefit of concept-based explanations is that they can aid the insertion of expert knowledge in the learning process of a model, allowing users to impose explicit domain-driven constraints defined as concepts, attributes, and predicates (for example, in so-called Logic Tensor Networks \cite{badreddine2022logic}). 
However, explanations based on human-understandable concepts are still in early development. In particular, concept-based explanations are mostly elaborated only for classification or regression models, leaving aside other problems and models in which concept-based explanations could be useful. This could be the case for reinforcement learning, in which explanations should inform about how the agent's interaction with \textit{concepts} existing in the environment produces a series of actions that fulfil the formulated task \cite{heuillet2021explainability}. 
Furthermore, limited work investigates \ac{XAI} methods that aim at synthesizing human-understandable concepts in concrete applications. While some concepts are universal, such as \enquote*{every car has tires and tires are round}, others are more subjective or differ among stakeholders and cultures and depend on domain knowledge, that is, knowledge related to training data \cite{mes21}. Accordingly, using a method that is generalisable and applicable across diverse areas and different contexts is needed, as one might be interested in using concepts in a personalized way to explain.

\paragraph*{Solution Ideas}
Creating concept-based \ac{XAI} requires a multi-faceted approach that takes into account a broad range of sub-problems. It begins with finding reliable ways for the extraction and identification of relevant concepts from data or \ac{AI} models.
For this first step, employing techniques from natural language processing, semantic analysis, and domain-specific knowledge can assist in systematically pinpointing concepts. This systematic identification lays the foundation for offering insights rooted in real-world, comprehensible terms.
Next, the concepts must be personalized so that they are tailored to the individual consuming them. Allowing users to define their own concepts would be one way to ensure personalization. Interdisciplinary collaboration and continuous feedback loops could refine these concepts, making them more meaningful and interpretable. A supplementary avenue could be to organize concepts within a hierarchical structure. This structure could be useful for delivering explanations that can be tailored to different levels of granularity. This hierarchy may allow users (or the \ac{XAI} methods) to select explanations that match specific needs.
Technical challenges include identifying and minimizing the inaccuracies of synthesized concept-based explanations, which could be tackled by introducing quality metrics for concept-based explanations. Finally, the application of \ac{XAI} methods based on concept synthesis in different domains and applications is another sub-problem. This might be solved by personalization, as described above.

\subsubsection{Addressing Explanations Divorced From Reality}
\label{sec:divorce}
The complexity of information flows in increasingly complex \ac{AI} systems can result in what we call a \enquote*{reality drift}. As \ac{AI} systems are becoming smarter, their decision-making is becoming more intricate. \ac{AI} systems might start using concepts that are impossible to convey to humans \cite{weller_transparency_2019, hamon2021impossible}. This means that the concepts humans use to understand the world might no longer suffice to describe reality in a meaningful and useful way \cite{flack_multiple_2012}. Consequently, the workings of such systems would become necessarily incomprehensible to us, and the utility of explanations, which are increasingly divorced from reality, may be questionable.
To bridge this gap, one might initially think that new concepts are needed that both humans and machines can use. However, there are differences in how humans and machines store and process information, making the success of this approach uncertain. In general, explanations provided by \ac{AI} systems may seem plausible to humans but could be detached from actual reality. This raises important questions about the usefulness of explainability in ensuring \ac{AI} safety, especially when dealing with highly complex \ac{AI} systems that are hard to decipher \cite{juric_ai_2020, brcic_impossibility_2023}.

\paragraph*{Solution Ideas}
To address the gap between explanations and reality, one potential solution involves engaging society and implementing regulations that ensure that someone can be held accountable for the performance of \ac{AI} systems, especially in critical situations\footnote{The EU AI Act: \url{https://artificialintelligenceact.eu/the-act/}}.
To achieve this, it is crucial to ensure that explanations are falsifiable (see \ref{sec:falsifiability}). Selecting explanation forms based on their falsifiability enables market and legal control over the types of \ac{AI} systems used.
Future systems should also tackle the uncertainty in modelling explanations by incorporating information from ontologies. There are three research directions to consider from here. The first direction explores the proof of the (non)existence of specific concept properties, such as gap size, robustness, simplicity, and estimability, to mention a few. The second direction focuses on developing adaptive ontology-generation methods to track evolving reality. These methods create adaptable and robust ontologies with computational properties that respect the limitations of human understanding. Basically, this approach would enhance the relevance of explainability in the context of reality drift. The third direction is sociological and deals with updating ontologies within society after adaptations.
In addition, when seeking adversarial robustness, it is preferable to establish protectorates at the highest possible level of abstraction in the ontology generation process for computational efficiency \cite{Krakauer2023}.
This comprehensive approach aims to improve the alignment of \ac{AI} explainability with the dynamic nature of real-world scenarios.

\subsubsection{Uncovering Causality for Actionable Explanations}
Causality is arguably among the most desired properties when constructing a model from data. In this regard, uncovering causal connections learned through a model via explanations is a fundamental hope associated with \ac{XAI} \cite{adadi2018peeking, carvalho2019mac, DoshiVelez2017Towards}. However, off-the-shelf posthoc \ac{XAI} methods fail to disentangle the correlation represented in the learned model from the causation between observed variables and predictions, making it questionable whether received explanations are suitable for guiding people's actions \cite{beckers2022causal}.
Explanations which are purely based on correlations can hinder decision-making when a model's outputs contain important information for action, for instance, the probability of failure of a production facility in industrial forecasting. Actionable and action-guiding explanations derived from causal models are needed in the real world, especially in scenarios where decisions may affect people.
To address this issue, counterfactual generation methods for \ac{ML} methods have garnered attention \cite{guidotti2022counterfactual}. Contrary to most \ac{XAI} approaches, counterfactuals attempt to answer why a black box model leads to a certain prediction by helping users understand what would need to change at its output to achieve a desired result \cite{chou2022counterfactuals}. In this answer, several desired properties should be met, namely, proximity, plausibility, sparsity, diversity, and feasibility \cite{guidotti2022counterfactual}. 
However, most works only regard a subset of these when producing counterfactuals, ignoring challenging issues. These include the provision of plausibility guarantees in highly complex data or generating diverse samples for largely parametric generative models prone to fall into single modalities. Furthermore, there are few causal approaches for \ac{XAI} since finding causal relationships from observational data is extremely difficult to achieve \cite{cinquini2022calime}.

\paragraph*{Solution Ideas}
To tackle the need for actionable explanations, technological advancements in \ac{AI}, such as large generative models, can open new opportunities in counterfactual explanations. One assumption is that such advancements can endow the produced counterfactuals with some of the desired properties for explanations such as proximity, plausibility, sparsity, diversity, and feasibility. This has been approached recently in \cite{del2022exploring}, where counterfactuals are produced by means of an optimization problem formulated over conditional GANs comprising three different objectives: one related to plausibility, another one to sparsity, and a third one that relates to feasibility. With initial explorations of diffusion-based counterfactuals being reported in recent research \cite{sanchez2022diffusion,augustin2022diffusion}, questions such as how to sample-efficiently diversify adversarial outputs produced by these models will be interesting.
Another direction worth exploring is how to connect causal graphs, relating each input of the model with its output, particularly in high-dimensional data. Most expert knowledge is represented in terms of entities and semantic relationships that inherently encode cause-effect links, as in knowledge bases. The goal in this context is to automatically construct causal graphs for models that do not necessarily operate on concepts or entities but rather on raw data. A potential solution is interfacing learning algorithms with symbolic knowledge about how the world behaves so that explanations for models grounded on such established causal links are endowed with the sought actionability.

\subsection{Adjusting XAI Methods and Explanations}
Another class of challenges in \ac{XAI} is related to adjusting explanations. With the diverse range of applications of \ac{AI} systems, \ac{XAI} methods have to produce explanations that fit diverse stakeholders, domains, and goals. However, there is not yet enough research addressing these concerns.

\subsubsection{Adjusting Explanations to Different Stakeholders}
An explanation can be required by many different kinds of stakeholders during the development, evaluation, and use of an \ac{AI} system \cite{Langer2021What}. Each stakeholder brings their own attitudes, preferences, aptitudes, abilities, and previous experiences that influence the kind of explanation they require. 
Designing and tailoring explanations that are appropriate for each of these stakeholder types, both in terms of \textit{content} as well as \textit{format and presentation}, is an ongoing challenge. For example, in the business context, the same objective facts must be explained and tailored to the stakeholders' respective interests and objectives. A business person is usually mostly interested in the bottom line impact of an \ac{AI} system, a technical person is interested in the process and the validity of implementation, and a financial person is interested in the cash flow. Adding to that mix, the different educational backgrounds and language used necessarily call for very different explanations for each of the three actors.

\paragraph*{Solution Ideas}
Future work should investigate new ways to enrich explanations semantically by combining different types of \ac{XAI} methods and utilizing additional information sources (for example, training data, ontologies, and other modalities). Ideas from personalizing \ac{DL} models \cite{sch21pe} and, more specifically, creating personalized explanations \cite{sch19p} can be helpful. Explanations could also be made interactive. Humans should be able to refine explanations through interaction, as recently advocated in the reinforcement learning community through reinforcement learning from human feedback \cite{zhu2023principled, bewley2021interpretable}.

\subsubsection{Adjusting Explanations to Different Domains}
The domain and context in which explanations are consumed are critical. For example, explanations for using a self-driving car must differ greatly from those in a clinical decision support system. Each domain brings different assumptions, environments, expectations, and stakes. In self-driving cars, the details about the passengers are not as important, but adherence to regulation is paramount. In contrast, in a clinical situation, the patient details are crucial, but regulation does not (directly) prescribe decisions.
Making each explanation universally applicable, precise, and compact means omitting many details that pertain to a domain, certainly sacrificing the explanation's effectiveness. Instead, we take the domain as indispensable and build on it. This makes meaningful explanations dependent on the domain on whose peculiarities and context they are built. In this line of thought, research is starting to emerge focused on distinguishing between high-stakes and low-stakes domains \cite{bunt_are_2012, rudin_stop_2019, Langer2021What}. However, the influence of the domain in which an \ac{AI} is used has not been fully explored.

\paragraph*{Solution Ideas}
Domain-specific explanation models should be developed to cater to the unique requirements of various application areas. These models should incorporate relevant knowledge, terminology, and context-specific reasoning to provide meaningful explanations. Furthermore, research efforts should prioritize the development of guidelines and standards for context-aware explanations. These guidelines would provide a structured approach for \ac{AI} developers to assess the use of and determine the most suitable explanation strategy.

\subsubsection{Adjusting Explanations to Different Goals}
Another fundamental challenge is to adjust explanations to what they should achieve when being presented to a stakeholder. For instance, data scientists might want to develop an accurate data-driven model; a regulator might want to assess the fairness of an \ac{AI}-assisted loan offer; or a loan applicant might want to know the reason behind a rejection \cite{hamon2021impossible, hamon2022bridging, Langer2021What}. An underlying assumption is that \ac{XAI} seeks to achieve these desiderata by improving the mental model that a stakeholder has of an underlying \ac{AI}-system \cite{doshi2017towards, gunning_xaiexplainable_2019, krajna_explainability_2022, Langer2021What}. However, the understanding required for each desideratum might differ, requiring tailored explanations.

\paragraph*{Solution Ideas}
Adjusting explanations to different goals might not be possible without factoring in the stakeholder that has these goals. Accordingly, one approach is to employ a stakeholder-centric explanation strategy, recognizing that different stakeholders have distinct goals and information needs. For data scientists aiming to improve model accuracy, explanations can focus on technical model details, feature importance, and model performance metrics. Regulators seeking to assess fairness may require explanations related to fairness metrics, compliance with regulations, and potential bias sources. Meanwhile, end-users, such as loan applicants, often require clear, user-friendly explanations for transparency regarding \ac{AI}-driven decisions, allowing them to understand the reasons behind outcomes. This stakeholder-specific tailoring ensures that the goals pursued with explainability are met effectively.

\subsection{Mitigating the Negative Impact of XAI}

Although \ac{XAI} has noble goals, it might also have negative impacts that need to be avoided or mitigated.

\subsubsection{Mitigating Failed Support by XAI}
In some domains, especially in the medical domain \cite{ghassemi2021false}, the ineffective support by \ac{XAI} can sometimes be harmful. 
This has been associated with the so-called \enquote*{white-box paradox} \cite{cabitza2022color, cabitza2023rams}, which urges not to take the value of the support delivered by \ac{XAI} systems for granted.
There are two possible cases: failed and misleading explanations. The first case might occur when the advice from an \ac{AI} system is correct, but the associated explanation fails to inform the decision maker positively. This can happen because the explanation is inappropriate or wrong, to appear faulty, irrelevant, or unclear to users \cite{cabitza2022color}. In this situation, users might not accept the correct advice because of inadequate explanations.
The second case is perhaps even worse and paradoxical; it occurs when the inference or advice of an \ac{AI} system is wrong, but the synthesized explanations have a sufficient persuasive force for convincing users that such advice is correct. In this situation, users are misled and thus potentially prone to mistakes \cite{bansal2021does}.

\paragraph*{Solution Ideas}
A first option would be to detect failure situations and label them appropriately and reliably. Then, one possible course of action would be not to provide users with any \ac{XAI} support if this is deemed detrimental or irrelevant in a given setting, as for instance, in radiological settings (see \cite{ghassemi2021false, cabitza2023rams}).
Another approach to mitigate failed support by \ac{XAI} is to challenge the \textit{oracular} conception of \ac{AI} support. This conception assumes that \ac{AI} outputs are judged based on moral categories like right and wrong, with \ac{AI}-generated explanations serving as aids to help humans determine whether to trust the outputs. However, \ac{AI} systems were originally conceived as \textit{generative} and \textit{persuasive} technologies \cite{natale2021deceitful}, not oracular ones. 
This oracular nature can be characterized as an \textit{alethic} nature, which assumes that machines can, and should, always state the truth \cite{cabitza2021err}. Relaxing the expectation of truthfulness is feasible, especially when dealing with probabilistic outputs or uncertainty estimates from \ac{AI} systems.
To this end, we could introduce a third type of explanation, namely a \textit{perorative} explanation, alongside the two traditional types of explanations provided by \ac{XAI} systems:  \textit{motivational} and \textit{justificative} explanations. In legal terms, peroration refers to the conclusion of a speech or argument, where a speaker summarizes their main points and seeks to persuade an audience of their position. 
By providing a set of possible explanations for different \ac{AI}-based inferences, including opposing and contradictory ones, \ac{XAI} systems enhance accountability among human decision-makers. 
This approach can be likened to a judicial process, where opposing parties present evidence and arguments before an impartial judge makes the final decision, offering a more balanced perspective than the oracular approach. \cite{cabitza2021need, cabitza2021err, miller2023explainable}

\subsubsection{Devising Criteria for the Falsifiability of Explanations}
\label{sec:falsifiability}

Explanations are often requested to clarify issues such as accountability \cite{weller_transparency_2019, abdul_trends_2018, mittelstadt2019explaining, Baum2022Responsibility}. However, explanations might be wrong. In such a case, parties that did not contribute to a mistake could be held accountable. Unfortunately, there is a lack of clarity regarding when an explanation is incorrect and under what conditions it becomes falsifiable.
Falsifiability is a critical element in introducing a commitment to the explanations provided by \ac{AI} systems and understanding the potential consequences that follow. Without clear criteria for falsifiability, benchmarks for the correctness of explanations cannot be established, and it becomes challenging to hold \ac{AI} practitioners accountable for the accuracy of their explanations. In some cases, practitioners may rely too heavily on intuition rather than rigorous methods regarding interpretability. Therefore, the question of establishing what are ground truths for explainability in benchmarks and how they were produced are open questions. As a more ambitious goal, we may ask about the discriminability between very differing plausible explanations and their ordering concerning quality and acceptability.

\paragraph*{Solution Ideas}
Establishing criteria for falsifiability in \ac{XAI} could draw inspiration from the philosophy of science and related research fields. One potential solution lies in adopting the Popperian notion of falsifiability as a cornerstone of empirical science that can serve as a guiding principle \cite{sep-popper}. Within this framework, \ac{XAI} could systematically integrate hypothesis testing and experimentation to subject explanations to rigorous empirical examination. In this line of thought, some researchers have advocated for a framework that promotes falsifiable research in the field of explainability, emphasizing the need for precision and rigour in evaluating and validating explanations \cite{Leavitt2020Towards}.
Additionally, insights from epistemology and cognitive science can inform the development of standardized protocols for evaluating the correctness of explanations, drawing parallels with how empirical claims in the sciences are subjected to rigorous scrutiny. Furthermore, interdisciplinary collaboration between computer scientists, philosophers of science, ethicists, and cognitive psychologists can facilitate the development of a comprehensive framework that incorporates not only empirical falsifiability but also ethical considerations and cognitive principles. By anchoring \ac{XAI} practices in well-established principles from the philosophy of science and related disciplines, it can pave the way for more robust, accountable, and scientifically grounded explanations within \ac{AI} systems.

\subsubsection{Securing Explanations from Being Abused by Malicious Human Agents}
Explainability is an important aspect of human coordination with machines \cite{dosilovic_explainable_2018, juric_ai_2020, krajna_explainability_2022, krajna_explainable_2022}. This is especially true in the near term, where AI systems may not be competent enough for autonomous adversarial behaviour. \ac{XAI} involves understanding how \ac{AI} systems arrive at their inferences, decisions or recommendations and providing a clear explanation of the logic and reasoning behind these outcomes. 
The effectiveness and adequacy of explainability as a tool for AI safety may be limited in certain scenarios \cite{murdoch_definitions_2019}. For instance, 
\ac{AI} systems in the hands of malicious human actors, 
can pose significant challenges to explainability \cite{juric_ai_2020} through manipulation and adversarial attacks. As an example, employers may systematically discriminate job applicants by using socially misaligned \ac{ML} models while serving borderline plausible explanations to avoid detection.

\paragraph*{Solution Ideas}
The need for discriminating between different explanations ties directly to the falsifiability of explanations (see \ref{sec:falsifiability}). Furthermore, Concept-based explanations could help combat adversarial attacks, especially a recent form of such attacks that aim to trick both humans and classifiers \cite{schn22concept}. For example, a malicious sample might be detectable by comparing a concept-based explanation of an adversarial sample with that of a non-adversarial sample. However, this is challenging because explanations can also be manipulated and used to trick or deceive \cite{sch22dec}. Similarly, another application context includes forensic analysis, which aims to understand the concepts learned by a classifier \cite{schn2020ai}. Concept-based explanations could also be helpful for reflective learning from data \cite{schn23ref}, which means classifiers can be improved through processing explanations during training.

\subsubsection{Securing Explanations from Being Abused by Malicious Superintelligent Agents}
Explainability is an important aspect of \ac{AI} safety. Many of the challenges, highlighted in the literature \cite{yampolskiy_unexplainability_2020,brcic_impossibility_2023} and here, already showcase fundamental limitations on the human ability to understand the behaviour of current \ac{AI} systems. However, assuming no constraints on their design or physical limitations, future \ac{AI} systems may become so competent that understanding them becomes fundamentally impossible. Exacerbating this issue, using formal verification to guarantee benign behaviour may not be viable due to unverifiability \cite{yampolskiy_what_2017}.
In independent domains where \ac{AI} agents are non-adversarial, these issues are not of much concern. At worst, we are in a situation where cooperative \ac{AI} agents work for us and tell us fairytales that make us content. However, when it comes to adversarial scenarios, the question is how much our assimilated explanatory concepts are adversarially robust through the existence of some computational protectorates that leave not many exposed loopholes.
As the complexity and capabilities of \ac{AI} agents increase, these agents may discover ways to deliberately fool people by exploiting the tension between the explanatory concepts that emerge from human capabilities, perception, and action and between those that complex agents can utilize. If such were the case, and humans rely only on explainability for safety, malignant gain by \ac{AI} agents could be unbounded \cite{brcic_impossibility_2023}. 

\paragraph*{Solution Ideas}
Explainability can be an effective tool in ensuring the safety of \ac{AI} systems, even in the long term, assuming that the problem of alignment between the technical capabilities of \ac{XAI} methods and their application and utility for humans, in reality, is solved (see \ref{sec:divorce}). 
The effectiveness and adequacy of explainability as a tool for \ac{AI} safety may be limited in certain scenarios \cite{murdoch_definitions_2019}. For this reason, explainability should be only one part of every safety toolkit as it has its strengths and limitations that need to be complemented in a portfolio of approaches. Work on building such a portfolio is welcome, as there is a growing need. This line of work is more long-term and can be solved only partially by constructive approaches. At the same time, the other part would be a restraint in building powerful super-intelligent systems without strong reasons to believe they are aligned with our values.

\subsection{Improving the Societal Impact of XAI}
Research on explainable \ac{AI} and the derived methods, models and techniques used to create real-world applications can impact society.

\subsubsection{Facilitating Originality Attribution of AI-Generated Data and Plagiarism Detection}
A special challenge for the explainability of novel generative models, which we think warrants to be mentioned separately, exists with respect to originality attribution and plagiarism detection of \ac{AI}-generated data. Concerning the problem of originality attribution, pieces of art produced by generative models have been recently taken to exhibit a similar level of creativity as humans. In particular, contests won with \ac{AI}-generated art have stirred controversy concerning the intellectual property of the output of a model learned from third-party data \cite{boutin2023diffusion, thorp2023chatgpt}. 
Likewise, plagiarism detection is becoming central in \acp{LLM} that excel across different domains, for example, with ChatGPT \cite{van2023chatgpt}). The massive usage of these models to produce apparently original textual content has disrupted the idea of plagiarism, as such content has been shown to evade mainstream tools for plagiarism detection easily. Thus, whether information biases the generative process, as, for example, with the prompt in a language-to-image stable diffusion model, is sufficiently original for intellectual property and author attribution claims remains an open question.

\paragraph*{Solution Ideas}
Regarding originality attribution, a solution is reformulating the concept of authorship in these models both from the technical point of view and from the legal and regulatory perspectives. 
Explainability should play a part in future regulation, as explanations could reveal which instances or parts of the modelled data distribution are relevant for a given synthesized output of the model. 
Solutions should be devoted to understanding if generalisation implies any form of plagiarism or whether it is a new form of inspiration, interfacing creative thoughts with original synthesized content.
On the topic of plagiarism detection, efforts have been made recently to determine whether the content produced by AI models is artificially generated, proposing the inclusion of tailored tokens, for example, \textit{watermarking} \cite{boenisch2021systematic}, in the produced content \cite{kirchenbauer2023watermark}. Explainability techniques will be relevant in determining which learning instances were more influential in producing a given outcome.

\subsubsection{Facilitating the Right to Be Forgotten}
Large-scale generative models require tons of data to fine-tune their trainable parameters, which often account for several hundreds of terabytes in size. Such a huge data substrate may clash with a fundamental right in data governance: the right to be ignored or forgotten by data-driven models.
While the interest in the \textit{machine unlearning} paradigm \cite{bourtoule2021machine} has been on the rise \cite{nguyen2022survey}, it is unclear how to efficiently ensure that data owned by a certain user is \textit{unlearned} by a given large-scale generative model, so that it can be ensured that no instance like that of the user claiming their right to be forgotten will be produced when the model is queried.

\paragraph*{Solution Ideas}
The right to be forgotten could be supported by similarity-based explanations and incrementally retraining the model to avoid sampling around the part of the subspaces close to the forbidden data. \ac{XAI} can also play a pivotal role by explaining model decisions and revealing which data points influenced those decisions. This transparency can empower users to identify the data instances that relate to them, enabling them to exercise their right to be forgotten. 
Additionally, \ac{XAI} can aid in auditing and verifying that the unlearning process is carried out effectively, reassuring users that their privacy rights are upheld. In general, humans should be granted the chance to verify that a generative model does not learn from them.

\subsubsection{Addressing the Power Imbalance Between Individuals and Companies}
A significant issue in \ac{XAI} is that the efforts in guaranteeing more transparency of \ac{AI} systems often are not enough to mitigate or even address the problem of unfair \ac{AI} systems that exacerbate the societal power imbalance between individuals and companies using \ac{AI} systems \cite{cohen2019between, cabitza2023quod, hamon2021impossible, malgieri2021just}. In other words, explaining the logic of an algorithm might be essential to empowering individuals to understand how to react to unreasonable \ac{AI}-driven systems, especially when those systems take automated decisions that can legally or similarly significantly affect individuals. Still, explainability is often hard to achieve in practice and limited in scope. 
The capability to understand \enquote*{why} a certain automated system followed a path from some inputs to some outputs may not be enough to empower individuals in case such a path was logically correct but legally or ethically disputable. Explanations are not enough if they are not accompanied by accountable systems of \textit{contestability} \cite{bayamlioglu2018contesting} and by justificatory statements that could prove why the \enquote*{path} from inputs to outputs is not only logically correct but also non-discriminatory, non-manipulative, non-illegal, non-unfair \cite{henin2021beyond, henin2021framework}.
Therefore, only acting at the level of the individual \enquote*{reactions} to the outputs of automated decision-making, including understandability, contestability, and justifiability, fails to completely address the main societal and ethical problems behind unfair and untrustworthy \ac{AI}. The \ac{XAI} community should shift its focus to tackle the power imbalance between \ac{AI} developers or controllers and those affected by \ac{AI} \cite{austin2014enough, wilsdon2022carissa}. 
The power imbalance is a structural problem, but the way \ac{AI} increases such an imbalance cannot be faced only by more explainability. There is a broader problem of under-representation, hidden discrimination, and lack of accountability \cite{costanza2020design}. Current \ac{XAI} methods answer this issue, but they can address only a small part of the problem \cite{kaminski2020algorithmic}.

\paragraph*{Solution Ideas}
To address the power imbalance between individuals and companies in the realm of \ac{XAI}, a new approach to designing future \ac{AI} systems via \ac{XAI} methods can include participative design, where impacted stakeholders are invited into the decision-making process \cite{costanza2020design, gregory2003scandinavian}. There are different modalities of participative approach to \ac{AI} design, but an essential consideration is the participative impact assessment \cite{mantelero2022beyond}. Vulnerable impacted stakeholders should be included, through an open and circular approach, in the key value-sensitive decisions in the \ac{AI} design \cite{malgieri2023acceptable}. 
Following the example of environmental impact assessment or workers' participation in business decision-making \cite{bodker2009participatory}, there are different ways in which digital users, individually, in groups or through representatives, can participate in the data processing decision-making or the design of data-driven technologies.

\section{A Novel Manifesto}\label{sec:novel Manifesto}
We conclude this article by presenting a manifesto for \ac{XAI}. This manifesto aims to define and succinctly describe the open challenges scholars in the field face. 
It includes propositions governing independent scientific research. The Manifesto is a mechanism for shaping our shared visions about science in the field of \ac{XAI}, and it is the outcome of the engagement of diverse expertise and different experiences by its authors.

\begin{enumerate}
  \item \textbf{Creating Explanations for New Types of \ac{AI}:} To create explanations for generative models (for example, \acp{LLM}) and for concept-based learning algorithms.   
  \item \textbf{Improving (and Augmenting) Current \ac{XAI} Methods}: To augment and improve attribution methods, remove artefacts in synthesis-based explanations, and create robust explanations. 
  \item \textbf{Evaluating \ac{XAI} Methods and Explanations:} To facilitate the human evaluation of explanations, create an evaluation framework for \ac{XAI} methods, and overcome limitations of studies with humans. 
  \item \textbf{Clarifying the Use of Concepts in \ac{XAI}:} To clarify the main concepts in \ac{XAI} and its relationship to trustworthiness, and to find a useful account of understanding. 
  \item \textbf{Supporting the Multi-Dimensionality of Explainability:} To create multi-faceted explanations and enable interdisciplinary work in \ac{XAI}.   
  \item \textbf{Supporting the Human-Centeredness of Explanations:} To create human-understandable explanations, facilitate explainability with concept-based explanations, address explanations divorced from reality, and uncover causality for actionable explanations.
  \item \textbf{Adjusting \ac{XAI} Methods and Explanations:} To adjust explanations to different stakeholders, domains, and goals.  
  \item \textbf{Mitigating the Negative Impact of \ac{XAI}:} To adjust explanations to different stakeholders, devise criteria for the falsifiability of explanations, and secure explanations from being abused by malicious human or superintelligent agents. 
  \item \textbf{Improving the Societal Impact of \ac{XAI}:} To facilitate the originality attribution of AI-generated data and plagiarism detection, support the right to be forgotten, and address the power imbalance between individuals and companies.
\end{enumerate}

\medskip
We believe working together as a community will lead to more productive and up-to-date work, increase reliability and enhance falsifiability.
The spirit of close collaboration, even among scholars with different scientific backgrounds and focused on specific disciplines, along with the respect and the willingness to build on each other's work, will certainly inspire more scholars to join us in advancing eXplainable Artificial Intelligence as a field. 
This manifesto is a genuine attempt at it, an exciting opportunity for shaping the future of AI-based systems for the benefit of human society.

\newpage
\bibliographystyle{unsrt}  
\bibliography{biblio}

\end{document}